\documentclass[10pt,twocolumn]{SICE19}
\usepackage{cite}
\usepackage{amsmath,amssymb,amsfonts}
\usepackage{algorithm}
\usepackage[noend]{algpseudocode}
\usepackage{graphicx}
\usepackage{textcomp}
\usepackage{xcolor}

\usepackage{bm}
\usepackage{dsfont}
\usepackage{svg}
\usepackage{amsmath}
\usepackage{amssymb}
\usepackage{flushend}

\def\BibTeX{{\rm B\kern-.05em{\sc i\kern-.025em b}\kern-.08em
    T\kern-.1667em\lower.7ex\hbox{E}\kern-.125emX}}

\title{State Estimation of Continuum Robots:\\ A Nonlinear Constrained Moving Horizon Approach}
\author{Hend Abdelaziz$^{*1\dagger}$, Ayman Nada$^{1}$, Hiroyuki Ishii$^{2}$, and Haitham El-Hussieny$^{1}$}
\speaker{Hend Abdelaziz}

 \affils{${}^{1}$Department of Mechatronics and Robotics Engineering, Egypt-Japan University of Science and Technology, Alexandria, Egypt.\\
 (E-mail: hend.abdelaziz@ejust.edu.eg, ayman.nada@ejust.edu.eg, and haitham.elhussieny@ejust.edu.eg)\\
 ${}^{2}$Faculty of Science and Engineering, Waseda University, Tokyo 169-8050, Japan.\\
 (E-mail: hiro.ishii@waseda.jp)\\
 }
\abstract{%
Continuum robots, made from flexible materials with continuous backbones, have several advantages over traditional rigid robots. Some of them are the ability to navigate through narrow or confined spaces, adapt to irregular or changing environments, and perform tasks in proximity to humans. However, one of the challenges in using continuum robots is the difficulty in accurately estimating their state, such as their tip position and curvature. This is due to the complexity of their kinematics and the inherent uncertainty in their measurement and control. This paper proposes a moving horizon estimation (MHE) approach for estimating the robot's state, including its tip position and shape parameters. Our approach involves minimizing the error between measurement samples from an IMU attached to the robot's tip and the estimated state along the estimation horizon using an inline optimization problem. We demonstrate the effectiveness of our approach through simulation and experimental results. Our approach can potentially improve the accuracy and robustness of state estimation and control for continuum robots. It can be applied to various applications such as surgery, manufacturing, and inspection.}
\begin{document}
\maketitle

\section{Introduction}

Soft and continuum robots, made of flexible materials such as silicone or rubber, are capable of continuous shape and movement changes \cite{robinson1999continuum, majidi2021review}. Their unique properties make them well-suited for a wide range of applications, making them safely interact with humans and the environment \cite{ashuri2020biomedical, el2020nonlinear}. Unlike rigid robots, which can be dangerous or destructive when colliding with delicate objects or individuals, soft and continuum robots can adapt to their surroundings and apply minimal force \cite{seleem2020development}. This makes them ideal for use in rehabilitation \cite{Wang2022DesignAR} and assistive care \cite{majidi2021review} scenarios, where robots must work closely with humans without causing harm. Additionally, the softness of these robots allows them to maneuver easily in narrow, confined, or challenging environments making them ideal for tasks such as exploration, monitoring, and cleaning\cite{milana2022soft, coad2019vine}.

Soft and continuum robots are known for their flexibility in navigating complex environments. However, accurate sensors are crucial for providing precise feedback on the robots' shape and position \cite{mahoney2016inseparable}. Estimating the state of continuum robots, including their configuration, shape, and location, is critical for various applications, especially since it is not always feasible to use sensors to acquire knowledge of full state due to financial and compactness considerations \cite{el2018development}. Accurate state estimation enables feedback control of the unknown state, simplifies control selection, and improves control accuracy \cite{rus2015design, yip2014model}. However, achieving accurate state estimation is challenging due to the noise and inaccuracies inherent in sensor readings \cite{mahoney2016inseparable}.


Various conventional state estimation methods have been employed to estimate the state of continuum robots. For example, Loo et al. \cite{loo2019non} proposed an Extended Kalman Filter (EKF) \cite{kalman1960contributions} framework to estimate the curvature angle of a pneumatic-based tentacle soft robot using nonlinear system identification and state estimation. In \cite{guo2019continuum}, a real-time method is proposed to estimate the shape change of a continuum robot using small permanent magnets and magnetic sensors. The method involves detecting the posture change of magnets and calculating the relative bending angle to reconstruct the shape in a 2D plane. Furthermore, in \cite{lilge2022continuum}, a Gaussian process regression approach is proposed for state estimation of continuum robots. This method aims to estimate the continuous shape and strain of the robot, given discrete and noisy measurements.

Although there have been advancements in state estimation techniques applied to estimate the state of soft continuum robots, the constraints imposed by the robot's workspace or configuration have not been fully considered. These constraints could arise from the robot's actuator limits, dimensions, or workspace boundaries. Furthermore, the non-linearity of the continuum robot's dynamics has not been fully exploited in the state estimation process, as linear models are continuously computed around the estimated state to apply linear state estimation techniques. In this paper, a Moving Horizon Estimation (MHE) \cite{ferrari2002moving} is utilized as a dynamic optimization approach to estimate the unknown state of continuum robots over a finite time horizon. The MHE has the ability to handle non-linear systems and constraints, cope with noise and uncertainty in measurements, and provide a real-time estimation of the system's state.

Our main contribution is the application of MHE for state estimation of soft continuum robots, which effectively handles the non-linearity of the robot's motion model. By incorporating a series of measurements from an IMU attached to the robot tip, our proposed MHE approach provides more accurate and reliable estimates of the robot's state compared to the conventional EKF method. Additionally, our approach can take into account the constraints imposed by the robot's workspace or configuration, which has not been fully explored in previous studies. The improved state estimation accuracy and ability to handle constraints make our proposed method more suitable for various applications. Overall, our work demonstrates the potential of MHE for soft continuum robot state estimation and opens up new avenues for future research in this direction.

The paper is organized as follows: Section \ref{sec:mhe} details the proposed MHE state estimation algorithm, including the motion and sensor models, cost function, constraints, and state estimation procedures. In Section \ref{sec:results}, we present the simulation and experimental results and compare the performance of our MHE approach with the Extended Kalman Filter (EKF) state estimation method. Finally, Section \ref{sec:conc} concludes the paper, summarizes our main contributions, and outlines potential directions for future work.

\section{Moving Horizon State Estimation}
\label{sec:mhe}
\begin{figure}[!b!p]
    \centering
    \includegraphics[width=0.9\columnwidth]{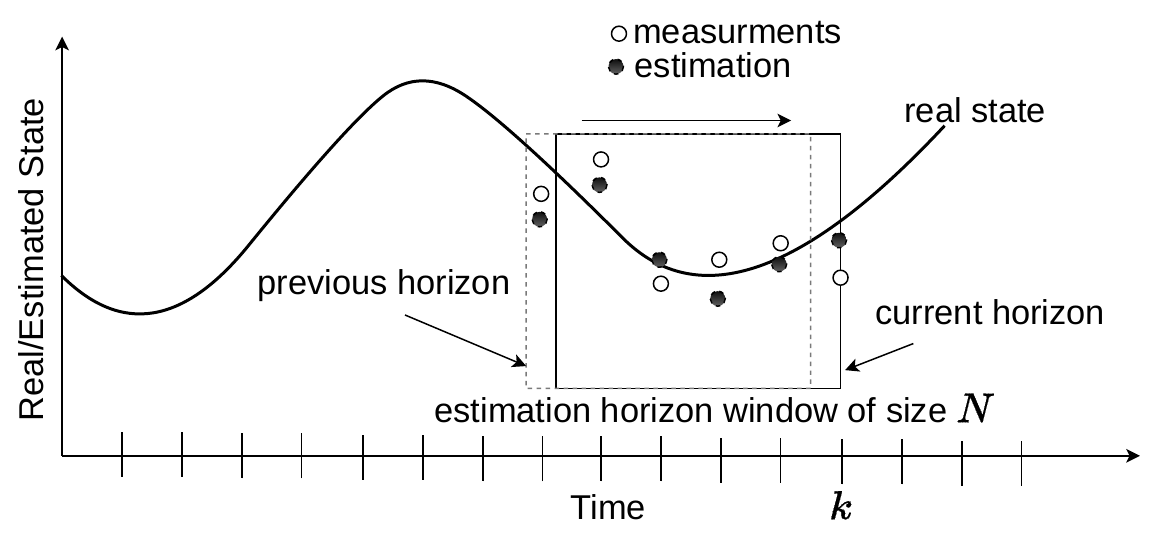}
    \caption{\label{fig:mhe_block}The fundamental idea of MHE involves utilizing an estimation horizon window with a fixed window size $N$.}  
\end{figure}

Accurately determining the state of a continuum robot, including its tip position and shape parameters, is crucial for successfully performing navigation and manipulation tasks. Soft continuum robots pose a challenge in that attaching numerous sensor nodes along the backbone of the robot can be cumbersome and expensive, leading to increased complexity in the control system. Therefore, utilizing a state estimation algorithm is a key approach to obtaining the robot's tip position in Cartesian space and its shape parameters. Moving Horizon Estimation (MHE) is a technique employed to acquire a complete estimation of the state of a continuum robot, which is crucial for applications that require knowledge of the robot's tip position, such as stiffness imaging. MHE is an optimization method that employs the robot's differential kinematic model as the motion model and the sensor's measurement model to accurately estimate the robot's state based on a window of sampled noisy measurements as shown in Fig. \ref{fig:mhe_block}. Additionally, MHE accounts for any constraints that may arise during the process, such as actuator limits or environmental settings, ensuring that the estimation remains feasible.

\subsection{Motion Model of Continuum Robots}

\begin{figure}[!b!p]
    \centering
    \includegraphics[width=.65\columnwidth]{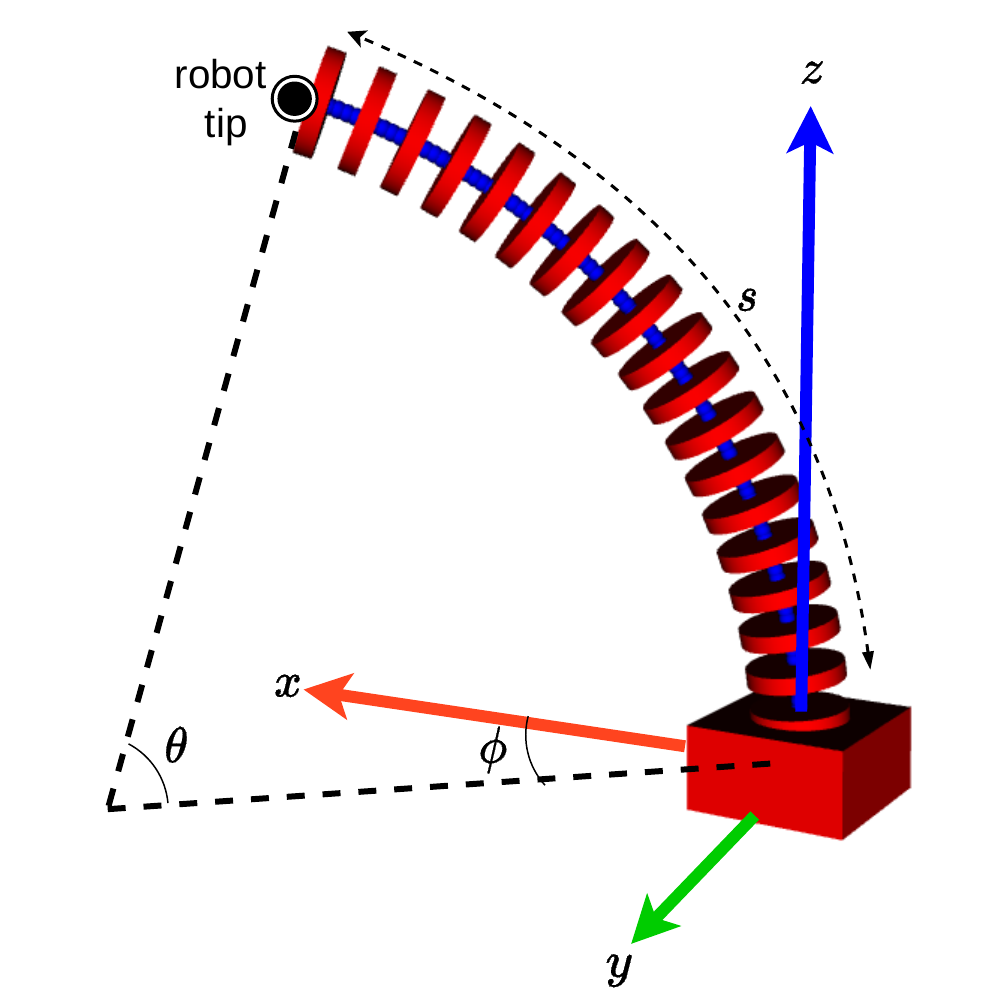}
    \caption{A one-section continuum robot with shape parameters: $s$, $\theta$, and $\phi$.}
    \label{fig:kinem}
\end{figure}

The MHE employs the robot's kinematic model to predict the evolution of the robot's state over a finite time horizon. The motion model captures the dynamics of the robot's motion and is used to propagate the state estimate over time. To develop a kinematic model for the continuum robot, we utilized the constant curvature assumption \cite{webster2010design}, which assumes that a one-section continuum robot is a segment of a circle with a constant radius. As illustrated in Fig. \ref{fig:kinem}, we considered the robot to be non-extensible, with a fixed length denoted by $s$. The position of the robot's tip is determined by: the angle of curvature, ${\theta}$, and the angle of curvature plane ${\phi}$ relative to the base's x-axis.
The homogeneous transformation matrix $\bm{T}_e^b\in \mathds{R}$$^{4 \times 4}$ that describes the relationship between the robot's tip and its base, as a function of the robot's shape parameters, is expressed as follows:
\begin{equation}
\label{eq:transf_matx}
     \bm{T}_e^b=
    \begin{bmatrix}
    \bm{R}_e^b  & \bm{p}_e^b\\
    \bm{0}_{1\times3}& 1
    \end{bmatrix}
\end{equation}
Here, $\bm{R}_e^b$ represents the rotation matrix and $\bm{p}_e^b$ denotes the translation vector, relating the orientation and position $[x, y, z]$ of the robot's tip to its base.
 
\begin{equation}
\label{eq:rot_mtx}
    {\bm{R}_e^b=
    \begin{bmatrix}
    C^{2}_{\phi}(C_{\theta}-1)+1 & S_{\phi} C_{\phi}(C_{\theta}-1) & -C_{\phi} S_{\theta}  \\
    S_{\phi} C_{\phi}(C_{\theta}-1) &  C^{2}_{\phi}(C_{\theta}-1)+C_{\theta} & 
    -S_{\phi} S_{\theta}\\
    C_{\phi} S_{\theta} & S_{\phi} S_{\theta} & C_{\theta} \\
    \end{bmatrix}}
\end{equation}

\begin{equation}
\label{eq:robot_pos}
    {\bm{p}_e^b= m(\bm{x}) =  \begin{bmatrix}
        x\\
        y\\
        z
    \end{bmatrix}=
    \begin{bmatrix}
    \dfrac{s\ C_{\phi}(C_{\theta}-1)}{{\theta}} \\
    \dfrac{s\ S_{\phi}(C_{\theta}-1)}{{\theta}} \\
    \dfrac{s\ S_{\theta}}{{\theta}} \\
    \end{bmatrix}}
\end{equation}
Here, $m(\bm{x})$ represent the kinematic model, and $S_{\{.\}}$ and $C_{\{.\}}$ represent the sine and cosine functions of an angle, respectively.  

 In order to incorporate the motion model of the robot into the MHE algorithm, it is necessary to express the robot's motion as a nonlinear function of the form:
$
\dot{\bm{x}} = f(\bm{x}, \bm{u})
$
where $\bm{x}=[x,y,z,\theta,\phi]^T$ denotes the state of the robot to be estimated, while $\bm{u} = [\dot{\theta}, \dot{\phi}]$ represents the velocities of the robot's shape parameters calculated from the difference between two consecutive readings over a known sample time. Therefore, by computing the velocity kinematics of the robot's tip position as given in Eq. \eqref{eq:robot_pos}, we obtain
\begin{equation}
\label{eq:jacob}
   \dot{x} = \begin{bmatrix}
       \bm{J}_{\theta, \phi}\ \bm{u}\\
       \bm{u}
   \end{bmatrix}
\end{equation}
 Here, $\bm{J}_{\theta, \phi} \in \mathds{R}$$^{3 \times 2}$ refers to the numerical Jacobian matrix, which is computed by differentiating Eq. \eqref{eq:robot_pos} with respect to the robot's shape parameters $\theta$ and $\phi$.

\subsection{Measurement Model}

\begin{figure}[!p!b]
\centering
\includegraphics[width=.7\columnwidth]{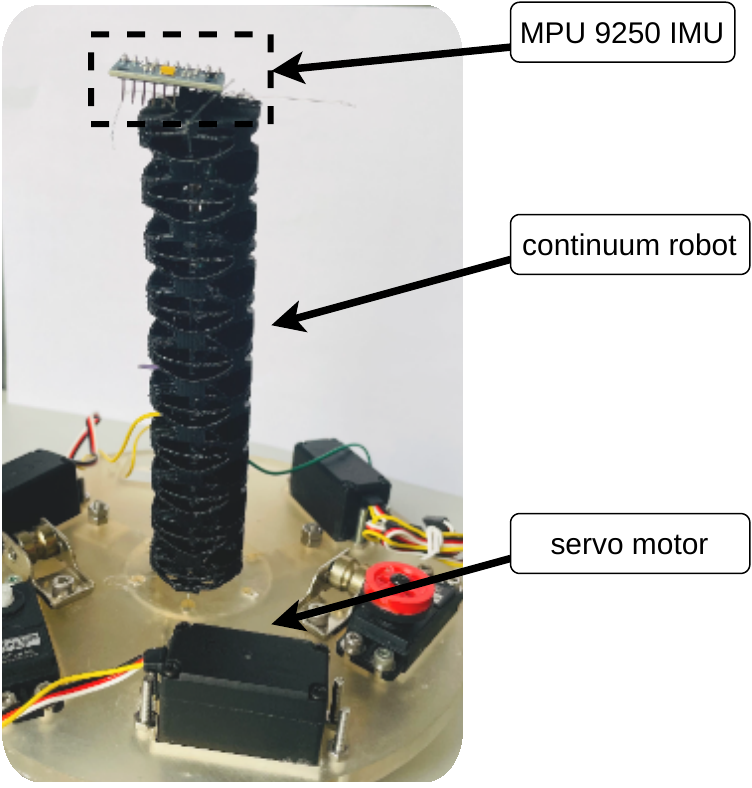}
\caption{\label{fig:imu} An MPU 9250 IMU sensor is attached to the tip of the one-section cable-driven continuum robot.}
\end{figure}

To correct the state estimate, the MHE algorithm utilizes the sensor's measurement model, which compares the predicted state of the robot with the actual measurements obtained from the sensors. This model captures the relationship between the robot's state $\bm{x}$ and the sensor measurements $\bm{z}$ and is represented by the function $\bm{z} = h(\bm{x})$. This function maps the robot's state to the expected sensor measurements, which are then compared to the actual sensor readings to update the state estimate. In this study, we employed an Inertial Measurement Unit (IMU) MPU9250 sensor, which was mounted on the tip of the continuum robot to measure its orientation as illustrated in Fig. \ref{fig:imu}. Specifically, we obtained the measurements $\bm{z} = [\gamma, \beta]^T$ as the roll ($\gamma$) and pitch ($\beta$) angles from the IMU sensor at each instance. The rotation matrix $\bm{R}(\gamma, \beta, \alpha)$ of the robot's tip in terms of the roll and pitch angles can be expressed as follows

\begin{equation}
\label{eq:rpy}
   {\scriptsize
  \bm{R}({\gamma},{\beta},{\alpha}) = 
    \begin{bmatrix} 
    C_{\alpha}C_{\beta} & C_{\alpha}S_{\beta}S_{\gamma}-S_{\alpha}C_{\gamma} & C_{\alpha}S_{\beta}C_{\gamma}+S_{\alpha}S_{\gamma} \\ 
    S_{\alpha}C_{\beta} & S_{\alpha}S_{\beta}S_{\gamma}-C_{\alpha}C_{\gamma} & S_{\alpha}S_{\beta}C_{\gamma}-C_{\alpha}S_{\gamma} \\ 
    -S_{\beta} & C_{\beta}S_{\gamma} & C_{\beta}C_{\gamma}
    \end{bmatrix}}
\end{equation}

By equating the rotation matrix $\bm{R}({\gamma},{\beta},{\alpha})$ in Eq. \eqref{eq:rpy} to the rotation matrix in Eq. \eqref{eq:rot_mtx}, we can obtain the expressions for the Euler roll and pitch angles as functions of the robot's state parameters:

\begin{equation}
\label{eq:measure}
\begin{bmatrix}
{\gamma} \\ {\beta}
\end{bmatrix}
=
\begin{bmatrix}
\sin^{-1}(-\cos{\phi}\sin{\theta}) \\
\tan^{-1}(\sin{\phi}\tan{\theta})
\end{bmatrix}
\end{equation}

These equations map the IMU measurements and the robot's estimated state, which is used in the MHE algorithm to correct the state estimate.
\subsection{Cost Function}
MHE utilizes a cost function that combines the motion and measurement models and captures the trade-off between prediction accuracy and measurement fidelity. The cost function determines the optimal state estimate, minimizing the prediction error and measurement noise. The MHE problem can be formulated as an optimization problem, with the states as the decision variables, that seeks to minimize the following cost given in 

\begin{equation} 
\label{eq:mhe_cost}
 J = \sum_{k=0}^{N-1} \Vert \bm{z}_k - h(\hat{\bm{x}}_k) \Vert^2_{\bm{V}} + \sum_{k=0}^{N} \Vert \bm{\hat{x}}_{k+1} - m(\bm{\hat{x}}_{k}) \Vert^2_{\bm{W}}
\end{equation}

where $\hat{\bm{x}}_k$ is the estimated state at instant k, and $\bm{V} \in \mathds{R}$$^{2 \times 2}$ and $\bm{W} \in \mathds{R}$$^{5 \times 5}$ are positive definite weighting diagonal matrices that represent the measurement noise and process noise, respectively. $N$ is the number of time steps defining the prediction horizon, which is a key parameter that determines the trade-off between the prediction accuracy and the computational complexity of the algorithm.

\subsection{Constraints}
MHE takes into account constraints that arise during the estimation process, such as actuator limits, the robot's dimensions, environmental settings, etc; which are incorporated into the MHE, ensuring the feasibility of the estimation. The MHE optimization problem given in Eq. \eqref{eq:mhe_cost} is subject to equality $g_1(\bm{x}, \bm{u}) = 0$ and inequality $g_2(\bm{x}, \bm{u})\leq 0$  constraints that defines the workspace of the robot.
The equality constraint ensures that the estimated state satisfies the motion model, $f(\bm{x},\bm{u})$, given in Eq. \eqref{eq:jacob} and is defined as:

\begin{equation}
 \label{eq:equality_constraints}
\bm{x}_{k+1} - \bm{x}_{k} + \Delta T\ f(\bm{x}_{k},\bm{u}_{k}) = \bm{0}, \quad \forall k \in [0,N]
\end{equation}

Meanwhile, a set of inequality constraints defining the workspace of the robot and bounding the velocity of the robot are given by

\begin{equation}
 \label{eq:inequality_constraints}
-s
	\leq
	\begin{bmatrix}
		x\\
  y\\
  z
  \end{bmatrix} \leq 
s, \quad \text{and} \quad
\begin{bmatrix}
    0\\-\pi
\end{bmatrix}	\leq	\begin{bmatrix}
		\theta\\
  \phi
  \end{bmatrix} \leq 
\begin{bmatrix}
    \pi/2\\ \pi
\end{bmatrix}
\end{equation}

It is important to note that all of the aforementioned constraints are considered as hard constraints, meaning that violating any of them will result in an infeasible solution and consequently an unsuccessful estimation.

\subsection{MHE State Estimation Algorithm}
The MHE algorithm for state estimation of the continuum robot in Algorithm \ref{alg:mhe} is presented by taking as input the motion model $f(.)$, measurement model $h(.)$, estimation horizon $N$, weighting matrices $\bm{V}$ and $\bm{W}$, and the initial $\bm{x}_0$ solution to the optimization problem. The output is the estimated state $\bm{x}_k$ of the robot for the given estimation horizon. The algorithm incorporates the constraints on the robot's workspace and shape parameters. The optimization problem is solved using the nonlinear solver Interior Point OPTimizer (IPOPT) \cite{dikin1967iterative}. 

\begin{algorithm}[h]

	\caption{\label{alg:mhe} MHE State Estimation Algorithm}
	\begin{algorithmic}[1]
		\Require
		\Statex Motion model, measurement model, and constraints: $f(\bm{x}, \bm{u}), h(\bm{x}), \bm{x}_0, g_1(\bm{x}), g_2(\bm{x}, \bm{u})$
		\Statex Weighting matrices $\bm{V}, \bm{W}$.
  		\Statex Estimation Horizon $N$.
    \Statex A series of  $M$ measurements $\bm{z}_k, \quad \forall k \in [0,M-1]$

		\Ensure
		\Statex The state estimation of the robot $\bm{x}_k, \quad \forall k \in [0,N]$.
		\Statex \hrulefill
		
		\State \textbf{For } $k=0$ to $M-1$:
		\State  \qquad $\bm{p} \gets \bm{z}_k, \forall k \in [k,k+N-1] $ \Comment{Get the measurements window}
		\State \qquad $\bm{x}_{k+1} \gets \text{IPOPT}(f, h, \bm{p}, g_1, g_2, \bm{x}_k)$ \Comment{Solve the optimization problem}
		\State  $\bm{x}_k \gets \bm{x}_{k+1},\ \forall k \in [k,k+N]$ \Comment{Estimation of the robot's state}
	\end{algorithmic}
\end{algorithm}

\section{Results and Discussion}
\label{sec:results}
This section presents the experimental results that evaluate the proposed MHE algorithm for estimating the state of the continuum robot.

\subsection{Performance Evaluation with Synthetic Data}
\begin{figure}[!p!b]
\centering
\includegraphics[width=0.85\columnwidth]{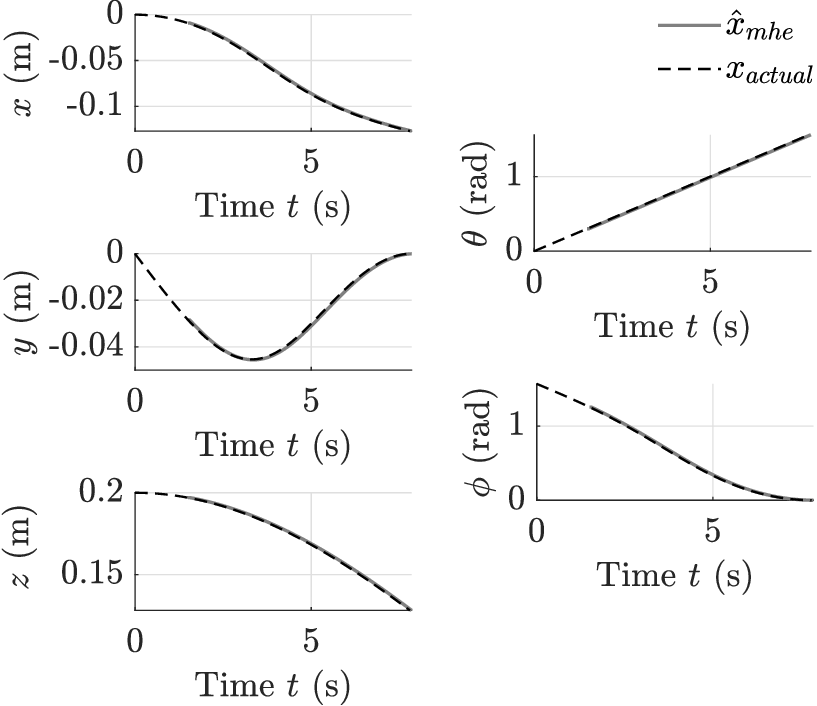}
\caption{\label{fig:mhe_3d} Results of proposed MHE state estimation using synthetic measurements.}
\end{figure}

First, we began by generating a set of synthetic measurements sampled from both angles, $\gamma \in [0, \pi/4]$ and $\beta \in [0, \pi/2]$. The synthetic data are generated such that the resulting actual state complies with the defined constraints. We used the measurement model described in Eq. \eqref{eq:measure} to obtain the corresponding shape parameters, $\theta$ and $\phi$, and finally, we obtained the actual state of the robot by solving the forward kinematics of the robot's tip, as presented in Eq. \eqref{eq:robot_pos}. The MHE algorithm was used to estimate the state of the robot for these measurements. A horizon window size of $N = 30$ was employed with a sampling time of $T = 0.05$ s, with weighting matrices chosen as
$$\bm{V} = \begin{bmatrix}
    2&0\\
    0&2
\end{bmatrix} \text{(degrees)}, \qquad \bm{W} = 10\bm{I}_{5 \times 5}$$

Fig. \ref{fig:mhe_3d} shows that the MHE algorithm provided a satisfactory estimation of the robot's state, with only minor discrepancies observed between the actual and estimated values.

\subsection{Constraints Handling}
\begin{figure}[!p!t]
\centering
\includegraphics[width=.85\columnwidth]{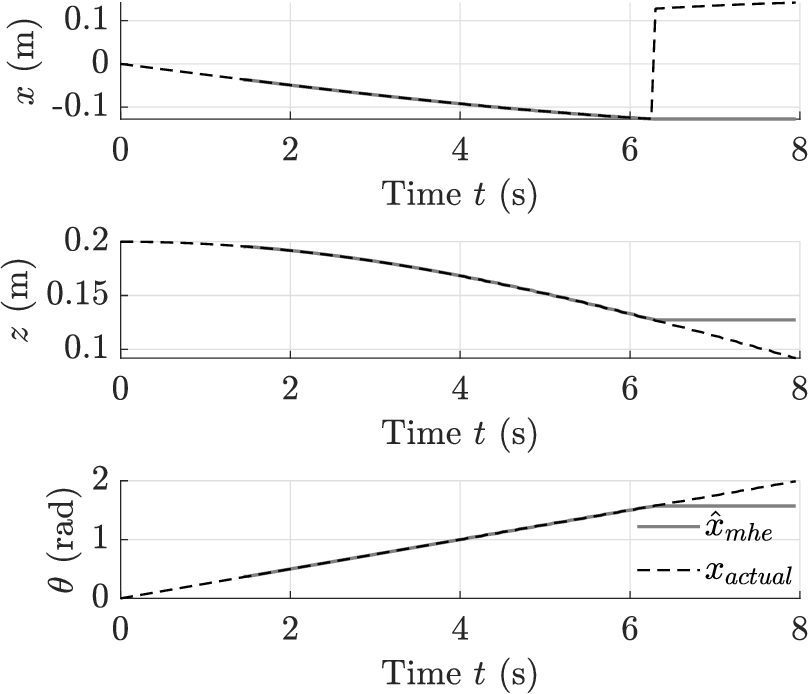}
\caption{\label{fig:mhe_2d} Results of MHE with synthetic two-dimensional data violating the robot's constraints.}
\end{figure}
Next, we evaluated the performance of the MHE algorithm with respect to the constraints imposed on the robot. We generated two-dimensional synthetic data that violate the workspace constraints. As shown in the results in Fig. \ref{fig:mhe_2d}, the MHE algorithm was able to estimate the robot's state with satisfactory performance until time $t \approx 6$ s, after which, the actual state violates the maximum allowable angle constraint, i.e., $\theta_{max} = \pi/2$. As demonstrated, the MHE algorithm respected the constraints and provided the maximum allowable estimation of the robot's state.

\subsection{Performance Evaluation with Real Data}
\begin{figure}[!p!b]
\centering
\includegraphics[width=\columnwidth]{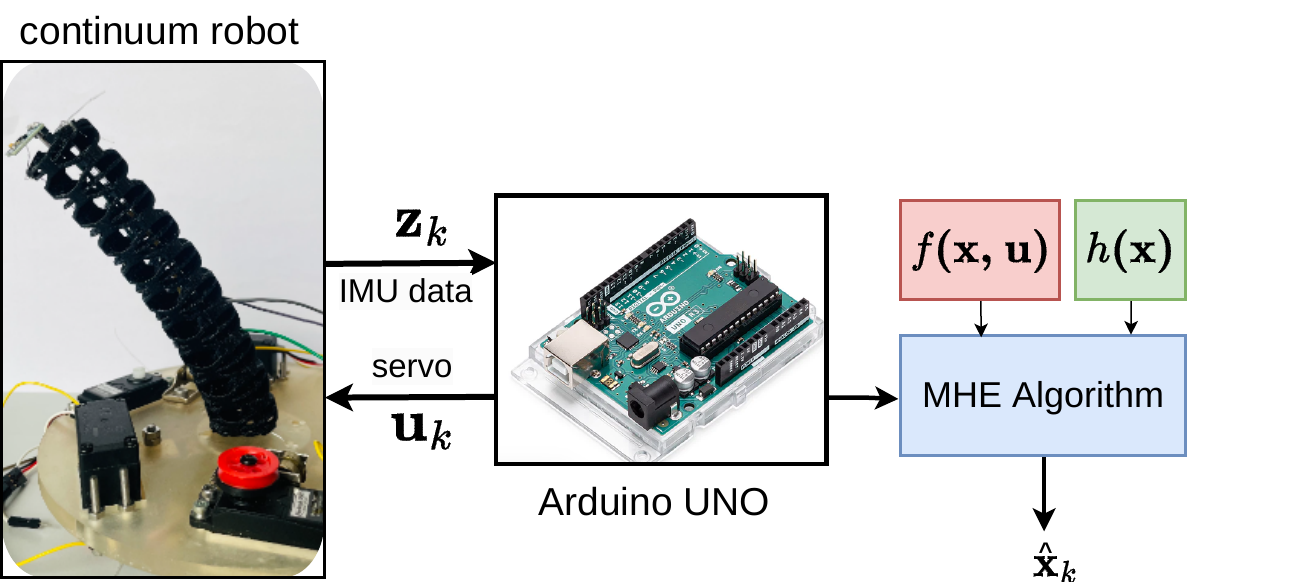}
\caption{\label{fig:robot} A block diagram showing the setup of the experiment of performance evaluation of the MHE for state estimation with real measurements.}
\end{figure}

\begin{figure}[!p!t]
\centering
\includegraphics[width=.85\columnwidth]{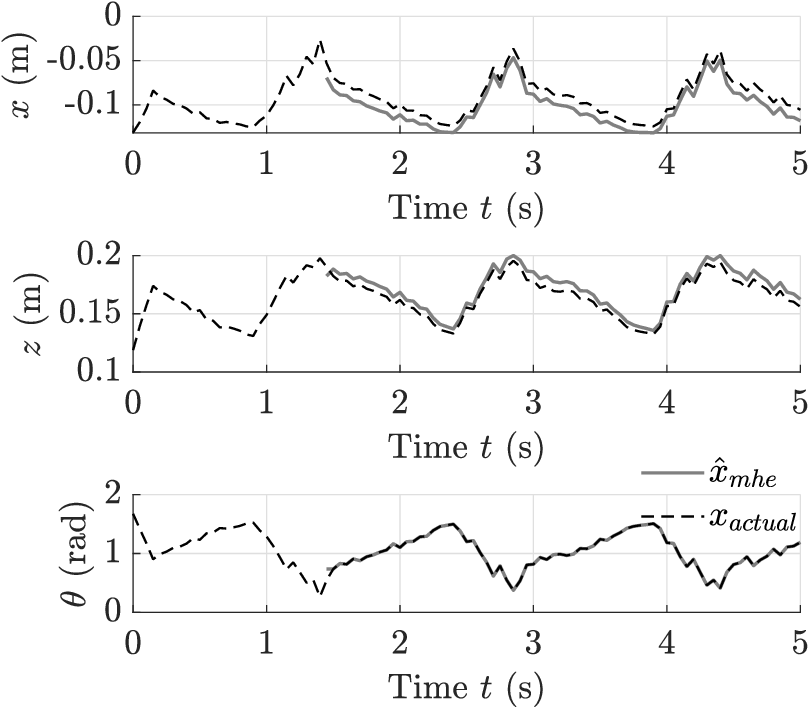}
\caption{\label{fig:mhe_motors} Results of applying the MHE algorithm to estimate the state of a one-section cable-driven continuum robot actuated by a single servo motor.}
\end{figure}

In this experiment, we evaluated the performance of the proposed MHE state estimation algorithm on measurements obtained from a physical one-section cable-driven continuum robot equipped with an attached IMU 9250 sensor, as shown in Fig. \ref{fig:robot}. The robot's tip was repetitively moved in a plane by actuating a single servo motor to apply tension to one robot's cable. An Arduino UNO was utilized to process the measurements and transmit them to MATLAB for applying the MHE state estimation. As shown in Fig. \ref{fig:mhe_motors}, the MHE algorithm demonstrates encouraging results in terms of following the actual trajectories obtained by solving the forward kinematics equation in Eq. \eqref{eq:robot_pos} using the real measurements. As shown, the MHE does not provide any estimations during the first $N$ samples because we require the previous $k-N$ measurements to estimate the state $\bm{x}_k$ at instance $k$. This highlights a possible area for improvement by proposing a dynamic window size $N_k$ instead of a fixed size horizon. This approach would enable the estimation of the robot's state more quickly with a smaller window size, which could then be increased to include more measurements and obtain a more accurate estimation.


\subsection{Comparison with Extended Kalman Filter (EKF)}
\begin{figure}[h]
\centering
\includegraphics[width=0.85\columnwidth]{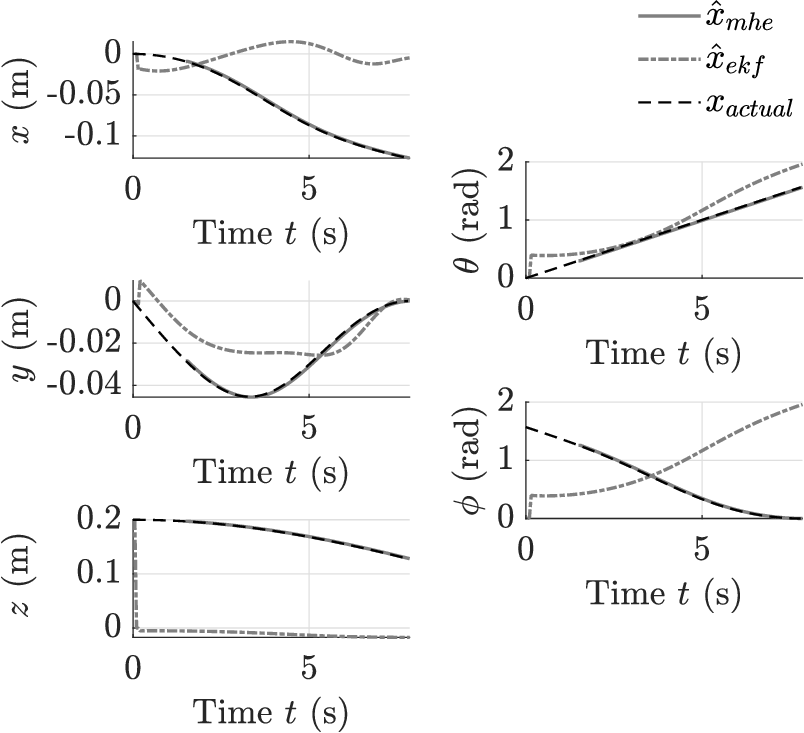}
\caption{\label{fig:mhe_vs_ekf_3d} Results of proposed MHE state estimation compared to that of EKF using synthetic measurements.}
\end{figure}

Fig. \ref{fig:mhe_vs_ekf_3d} illustrates the superior performance of our proposed MHE algorithm compared to the Extended Kalman Filter (EKF) in accurately estimating the state of continuum robots. The MHE's ability to fully handle the system's nonlinearity by incorporating a window of measurements and estimations leads to more accurate state estimation compared to EKF, which linearize the motion model at each estimation sample.

\begin{figure}[!p!t]
\centering
\includegraphics[width=.8\columnwidth]{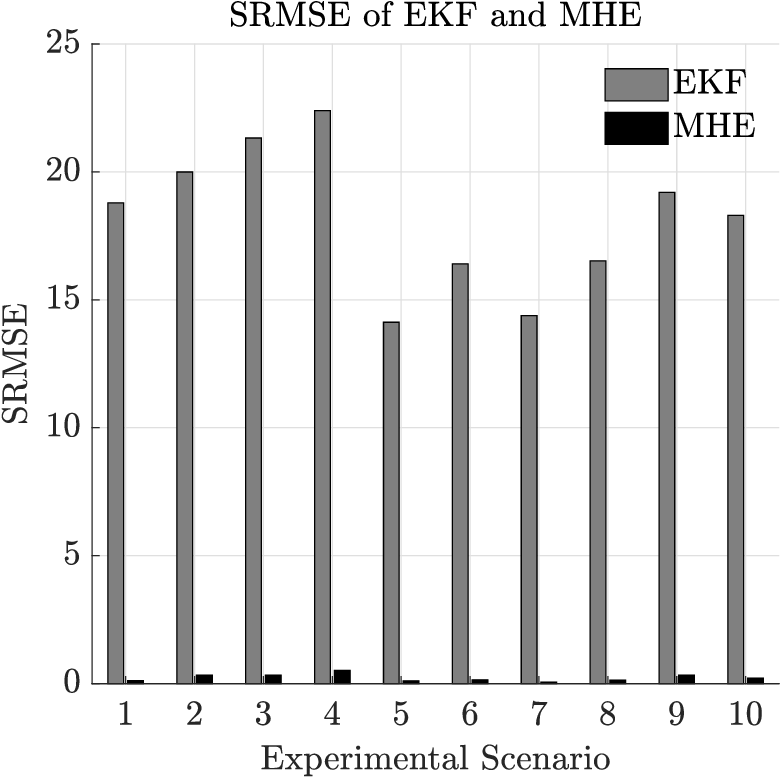}
\caption{\label{fig:mc} Results of the Monte-Carlo analysis conducted to evaluate the robustness and performance of the MHE and EKF against varying noise levels.}
\end{figure}

In addition, a Monte Carlo simulation was used to compare the robustness of the proposed MHE state estimation algorithm with the Extended Kalman Filter (EKF) algorithm under varying levels of measurement noise. We conducted ten simulation experiments with varying levels of noise added to both angles $\beta$ and $\gamma$, as described in Table \ref{table:montecarlo}, to compare the robustness of the proposed MHE with the EKF algorithm. The noise levels were added to the synthetic measurements, and for each scenario, we computed the Sum of Root Mean Square Error (SMRSE) for both algorithms, as depicted in Fig. \ref{fig:mc}. The results demonstrate that the proposed MHE algorithm is robust to variations in noise levels, whereas the EKF state estimation algorithm exhibits significant SRMSE values and is highly dependent on the level of noise. The robustness of the MHE algorithm can be attributed to its philosophy of incorporating a window of measurements into the state estimation process. This allows any noise added to the measurements to be averaged within the horizon window, contributing to the algorithm's robustness. In contrast, the EKF algorithm achieves a single-shot estimation of the robot state based on the current measurement and motion model, making it more susceptible to noise.

\begin{table*}[!p!t]
\caption{\label{table:montecarlo} Noise levels added to the angles $\beta$ and $\gamma$ in each experimental scenario.}
\centering
\begin{tabular}{|c|c|c|c|c|c|c|c|c|c|c|}
\hline
Scenario  & 1        & 2        & 3 & 4 & 5 & 6 & 7 & 8 & 9 & 10 \\ 
\hline
$\sigma_\beta$ (rad$^2$)  & 0.022 & 0.028  & -0.026   & 0.029   & 0.010    & -0.028    & -0.015   & 0.003   & 0.032    & 0.033    \\ \hline
 $\sigma_\gamma$ (rad$^2$) & -0.024  & 0.033 & 0.032   & -0.001   & 0.021   & -0.025     & -0.005   & 0.029   & 0.020     & 0.032    \\ \hline
\end{tabular}
\end{table*}

\subsection{Variation of Estimation Horizon}

\begin{figure}[!p!t]
\centering
\includegraphics[width=\columnwidth]{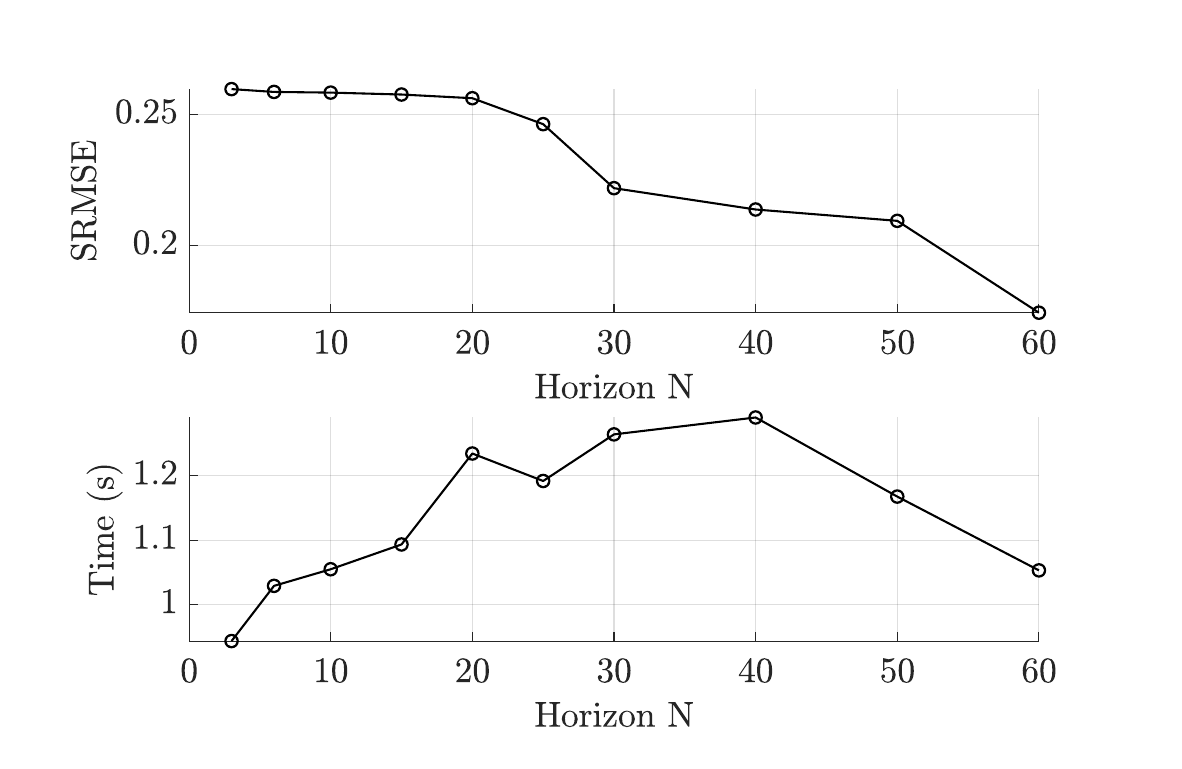}
\caption{\label{fig:est_hor} Effect of varying the estimation horizon $N$ on the SRMSE and the execution time.}
\end{figure}

Fig. \ref{fig:est_hor} shows the impact of increasing the estimation horizon $N$ on the estimation performance and mean computation time of the MHE algorithm. As shown, increasing $N$, at least up to a value of 40, results in a significant increase in the mean computation time. However, the SRMSE does not show a significant improvement with increasing $N$. The mean computation time increases rapidly until $N=40$ and then starts to decrease due to the fact that increasing the horizon length $N$ reduces the number of measurements used in the MHE state estimation, which is $M-N$ according to Algorithm \ref{alg:mhe}.
It is worth mentioning that increasing the estimation horizon $N$ could negatively affect the performance, particularly when it comes to incorporating the MHE state estimation into a real-time feedback control system. The controller has to wait for the first $N$ samples to estimate an unknown state reliably.

\section{Conclusion}
\label{sec:conc}
In conclusion, the proposed MHE state estimation algorithm has been successfully applied to estimate the state of a one-section continuum robot using both synthetic and real measurements from an IMU sensor attached to its tip. The algorithm showed promising results in accurately estimating the robot's shape and tip position, even when the actual state was out of the robot's workspace constraints compared to the EKF state estimation. Meanwhile, the proposed MHE algorithm demonstrated robustness in terms of its insensitivity to the change in the noise levels added to the measurements, as observed in the Monte-Carlo experiments conducted with varying noise levels. Furthermore, the results indicate the potential for dynamic window size adaptation in the MHE algorithm to improve estimation accuracy and speed. Overall, the MHE algorithm offers an efficient and effective approach to state estimation in continuum robots, which can be useful in various applications requiring high precision and accuracy.

\bibliographystyle{ieeetr}

\end{document}